# Tractable Inference for Complex Stochastic Processes

Xavier Boyen
Stanford University
*xb@cs.stanford.edu*

Daphne Koller
Stanford University
*koller@cs.stanford.edu*

## Abstract

The monitoring and control of any dynamic system depends crucially on the ability to reason about its current status and its future trajectory. In the case of a stochastic system, these tasks typically involve the use of a *belief state*—a probability distribution over the state of the process at a given point in time. Unfortunately, the state spaces of complex processes are very large, making an explicit representation of a belief state intractable. Even in dynamic Bayesian networks (DBNs), where the process itself can be represented compactly, the representation of the belief state is intractable. We investigate the idea of maintaining a compact approximation to the true belief state, and analyze the conditions under which the errors due to the approximations taken over the lifetime of the process do not accumulate to make our answers completely irrelevant. We show that the error in a belief state *contracts* exponentially as the process evolves. Thus, even with multiple approximations, the error in our process remains bounded indefinitely. We show how the additional structure of a DBN can be used to design our approximation scheme, improving its performance significantly. We demonstrate the applicability of our ideas in the context of a monitoring task, showing that orders of magnitude faster inference can be achieved with only a small degradation in accuracy.

## 1 Introduction

The ability to model and reason about stochastic processes is fundamental to many applications [6, 8, 3, 14]. A number of formal models have been developed for describing situations of this type, including Hidden Markov Models [15], Kalman Filters [9], and Dynamic Bayesian Networks [4]. These very different models all share the same underlying *Markov assumption*, the fact that the future is conditionally independent of the past given the current state. Since the domain is stochastic and partially observable, the true state of the process is rarely known with certainty. However, most reasoning tasks can be performed by using a *belief state*, which is a probability distribution over the state of a system at a given time [1]. It follows from the Markov assumption that the belief state at time $t$ completely captures all of our information about the past. In particular, it suffices both for predicting the probabilities of future trajectories of the system, and for making optimal decisions about our actions.

Consider, for example, the task of monitoring an evolving system. Given a belief state at time $t$ which summarizes all of our evidence so far, we can generate a belief state for time $t+1$ using a straightforward procedure: We propagate our current belief state through the *state evolution model*, resulting in a distribution over states at time $t+1$, and then condition that distribution on the observations obtained at time $t+1$, getting our new belief state.

The effectiveness of this procedure (as well as of many others) depends crucially on the representation of the belief state. Certain types of systems, e.g., Gaussian processes, admit a compact representation of the belief state and an effective update process (via *Kalman filtering* [9]). However, in other cases matters are not so simple. Consider, for example, a stochastic system represented as a dynamic Bayesian network (DBN). A DBN, like a Bayesian network, allows a decomposed representation of the state via state variables, and a compact representation of the probabilistic model by utilizing conditional independence assumptions. Here, a belief state is a distribution over some subset of the state variables at time $t$. In general, not all of the variables at time $t$ must participate in the belief state [3]; however, (at least) every variable whose value at time $t$ directly affects its value at time $t+1$ must be included. In large DBNs, the obvious representation of a belief state (as a flat distribution over its state space) is therefore typically infeasible, particularly in time-critical applications.

However, our experience with Bayesian networks has led us to the tacit belief that structure is usually synonymous with easy inference. Thus, we may expect that, here also, the structure of the model would support a decomposed representation of the distribution, and thereby much more effective inference. Unfortunately, this hope turns out to be unfounded. Consider the DBN of Figure 1(a), and assume to begin with that the evidence variable did not exist. At the initial time slice, all of the variables start out being independent. Furthermore, there are only a few connections between one variable and another. Nevertheless, at time 3, all the variables have become fully correlated: one can see from the unrolled DBN of Figure 1(b) that no conditional independence relation holds among any of them. Observing the evidence variable makes things even worse: in this case it takes only 2 time slices for all the state variables to become fully correlated. In general, unless a process decomposes into completely independent subprocesses, the belief state will become fully correlated very early in time. As any factored decomposition of a distribution rests on some form of conditional independence, no decomposition of this belief state is possible. This phenomenon is perhaps the most serious impediment to applying probabilistic reasoning methods to dynamic systems.

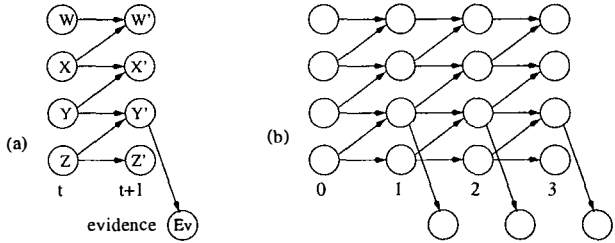

Figure 1: Example structured stochastic process: (a) two-time-slice DBN; (b) unrolled BN for 3 time slices.

Related problems occur in other types of processes. For example, in hybrid processes involving both continuous and discrete variables, we might wish to represent the belief state as a mixture of Gaussians. However, in most cases, the number of components in the mixture will grow exponentially over time, making it impossible to represent or reason with an exact belief state.

One approach to dealing with this problem is to restrict the complexity of our belief state representation, allowing inference algorithms to operate on it effectively. Thus, rather than using an exact belief state which is very complex, we use a compactly represented *approximate* belief state. For example, in the context of a DBN, we might choose to represent an approximate belief state using a factored representation. (See [12] for some discussion of possible belief state representations for DBNs.) In the context of a hybrid process, we might choose to restrict the number of components in our Gaussian mixture.

This idea immediately suggests a new scheme for monitoring a stochastic process: We first decide on some computationally tractable representation for an approximate belief state (e.g., one which decomposes into independent factors). Then we propagate the approximate belief state at time $t$ through the transition model and condition it on our evidence at time $t + 1$. In general, the resulting belief state for time $t + 1$ will not fall into the class which we have chosen to maintain. We therefore approximate it using one that does, and propagate further.

This general strategy must face the risk that our repeated approximations cause errors to build up out of control over extended periods of time, either by accumulation due to the repeated approximations, or (worse) by spontaneous amplification due to some sort of instability. In this paper, we show that this problem does not occur: the mere stochasticity of the process serves to attenuate the effects of errors over time, fast enough to prevent the accumulated error from growing unboundedly. More specifically, we have proved the important (and to our knowledge) new result that stochastic processes (under certain assumptions) are a *contraction* for KL-divergence [2]: propagation of two distributions through a stochastic transition model results in a constant factor reduction of the KL-divergence between them. We believe that this result will have significant applications in many other contexts.

This result applies to any discrete stochastic process and any approximate representation of the belief state. However, we show that even stronger results can be obtained in the case of structured processes equipped with an approximation scheme that is tailored to the structure of the process. Specifically, we consider processes that are composed of several weakly interacting subprocesses, and an approximation scheme that decomposes each belief state as a product of independent belief states over the individual processes. Under these assumptions, our contraction rate improves dramatically. We also show a strong connection between the amount of interaction among the subprocesses and the quality of our bounds. We provide empirical evidence for the success of our approach on two practical DBNs, showing that, by exploiting structure, we can achieve orders of magnitude faster inference with only a tiny degradation in accuracy. As we mentioned above, DBNs have been a counter-example to our intuition that structured representations leads to effective inference. Our results are the first to show how, at least for approximate inference, we can exploit the structure of a DBN for computational gain.

## 2  Basic framework

Our focus in this paper is on discrete-time finite-state Markov processes. Such processes can be modeled explicitly, as a *Hidden Markov Model* or, if additional structure is present, more compactly as a *Dynamic Bayesian Network*. A discrete time Markov process evolves by moving from one state to the other at consecutive times points. We use $S^{(t)}$ with $S^{(t)} \in S = \{s_1, ..., s_n\}$ to denote the state at time $t$. In the case of a DBN, $S^{(t)}$ may be described as an assignment of values to some set of state variables. The *Markov assumption*, inherent to all of the models we consider, asserts that the present state of the system contains enough information to make its future independent from its past, i.e., that $P[S^{(t+1)} \mid S^{(0)}, ..., S^{(t)}] = P[S^{(t+1)} \mid S^{(t)}]$. For simplicity, we also assume that the process is *time-invariant*, i.e., that the probability with which we have a transition from some state $s_i$ at time $t$ to another state $s_j$ at time $t + 1$ does not depend on $t$. Thus, we obtain that the process can be described via a *transition model* $\mathcal{T}$: $\mathcal{T}[s_i \leadsto s_j] \triangleq P[s_j^{(t+1)} \mid s_i^{(t)}]$, where we use $s_i^{(t)}$ to denote the event $S^{(t)} = s_i$. In the case of an HMM, $\mathcal{T}$ is often specified explicitly as an $n \times n$ matrix; for a DBN, $\mathcal{T}$ is described more compactly as a fragment of a Bayesian network (see Section 5.2).

The Markov process is typically hidden, or *partially observable*, meaning that its state is not directly observable. Rather, we observe a *response* $R^{(t)} \in R = \{r_1, ..., r_m\}$; in the case of a DBN, $R^{(t)}$ can be an assignment to some set of observable random variables. The response depends stochastically and exclusively on the state $S^{(t)}$; i.e., $R^{(t)}$ is conditionally independent of any $S^{(t')}$ and $R^{(t')}$ given $S^{(t)}$. Using $r_h^{(t)}$ to denote $R^{(t)} = r_h$, we obtain that the observability aspect of the process can be described via an *observation model* $\mathcal{O}$: $\mathcal{O}[s_i \hookrightarrow r_h] \triangleq P[r_h^{(t)} \mid s_i^{(t)}]$.

The Markov assumption implies that all the historical information needed to monitor or predict the system's evo-lution is contained in (the available knowledge about) its present state. This knowledge can be summarized in a *belief state*—a probability distribution over the possible states. We distinguish between the *prior* and the *posterior* belief state:

**Definition 1** The *prior belief state* at time $t$, denoted $\sigma^{(\bullet t)}$, is the distribution over the state at $t$ given the response history up to but not including time $t$. Letting $r_{h_k}^{(k)}$ denote the response observed at time $k$,

$$\sigma^{(\bullet t)}[s_i] \triangleq P[s_i^{(t)} \mid r_{h_0}^{(0)}, ..., r_{h_{t-1}}^{(t-1)}].$$

The *posterior belief state* at time $t$, denoted $\sigma^{(t\bullet)}$, is the distribution over the state at time $t$ given the response history up to and including time $t$:

$$\sigma^{(t\bullet)}[s_i] \triangleq P[s_i^{(t)} \mid r_{h_0}^{(0)}, ..., r_{h_{t-1}}^{(t-1)}, r_{h_t}^{(t)}]. \quad \square$$

The *monitoring* task is defined as the task of maintaining a belief state as time advances and new responses are observed. In principle, the procedure is quite straightforward. Assume we have a posterior belief state $\sigma^{(t\bullet)}$ at time $t$. Upon observing the response $r_h$ at time $t+1$, the new state distribution $\sigma^{(t+1\bullet)}$ can be obtained via a two-stage computation, based on the two models $\mathcal{T}$ and $\mathcal{O}$. The prior belief state $\sigma^{(\bullet t+1)}$ for the next time slice is obtained by propagating $\sigma^{(t\bullet)}$ through the stochastic transition model, while the posterior $\sigma^{(t+1\bullet)}$ is obtained by conditioning $\sigma^{(\bullet t+1)}$ on the response $r_h$ observed at time $t+1$:

$$\sigma^{(\bullet t+1)}[s_j] = \sum_{i=1}^{n} \sigma^{(t\bullet)}[s_i] \mathcal{T}[s_i \leadsto s_j],$$

$$\sigma^{(t+1\bullet)}[s_i] = \frac{\sigma^{(\bullet t+1)}[s_i] \mathcal{O}[s_i \hookrightarrow r_h]}{\sum_{l=1}^{n} \sigma^{(\bullet t+1)}[s_l] \mathcal{O}[s_l \hookrightarrow r_h]}.$$

Abstractly, we view $\mathcal{T}[\cdot]$ as a function mapping $\sigma^{(t\bullet)}$ to $\sigma^{(\bullet t+1)}$, and define $\mathcal{O}_{r_h}[\cdot]$ as the function mapping $\sigma^{(\bullet t+1)}$ to $\sigma^{(t+1\bullet)}$ upon observing the response $r_h$ at time $t+1$.

While exact monitoring is simple in principle, it can be quite costly. As we mentioned in the introduction, the belief state for a process represented compactly as a DBN is typically exponential in the number of state variables; it is therefore impractical in general even to feasibly store the belief state, far less to propagate it through the various update procedures described above.

Thus, we are interested in utilizing compactly represented approximate belief states in our inference algorithm. The risks associated with this idea are clear: the errors induced by our approximations may accumulate to make the results of our inference completely irrelevant. As we show in the next two sections, the stochasticity of the process prevents this problem from occurring.

## 3 Simple contraction

Consider the exact and estimated belief states $\sigma^{(t\bullet)}$ and $\hat{\sigma}^{(t\bullet)}$. Intuitively, as we propagate each of them through the transition model it "forgets" some of its information; and as the two distributions forget about their differences, they become closer to each other. As we will see in Section 5, in order for our errors to remain bounded, their effect needs to dampen exponentially quickly. That is, we need to show that $\mathcal{T}$ reduces the distance between two belief states $\sigma^{(t\bullet)}$ and $\hat{\sigma}^{(t\bullet)}$ by a constant factor.

This result is known for $\mathcal{T}$ if we use $L_2$ norm (Euclidean distance) as the distance between our distributions: $\| \mathcal{T}[\sigma^{(t\bullet)}] - \mathcal{T}[\hat{\sigma}^{(t\bullet)}] \|_2 \leq |\lambda_2| \cdot \| \sigma^{(t\bullet)} - \hat{\sigma}^{(t\bullet)} \|_2$, where $\lambda_2$ is the second largest eigenvalue of $\mathcal{T}$. Unfortunately, $L_2$ norm is inappropriate for our purposes. Recall that there are two main operations involved in updating a belief state: propagation through the transition model, and conditioning on an observation. Unfortunately, $L_2$ norm behaves badly with respect to conditioning: $\| \mathcal{O}_{r_h}[\sigma^{(\bullet t)}] - \mathcal{O}_{r_h}[\hat{\sigma}^{(t\bullet)}] \|_2$ can be arbitrarily larger than $\| \sigma^{(\bullet t)} - \hat{\sigma}^{(\bullet t)} \|_2$. In fact, one can construct examples where the observation of any response $r_h$ will cause the $L_2$ distance to grow.

Thus, we must search for an alternative distance measure for which to try and prove our contraction result. The obvious candidate is *relative entropy*, or *KL divergence*, which quantifies the loss or inefficiency incurred by using distribution $\psi$ when the true distribution is $\varphi$ [2, p.18]:

**Definition 2** If $\varphi$ and $\psi$ are two distributions over the same space $\Omega$, the *relative entropy* of $\varphi$ to $\psi$ is

$$D[\varphi \| \psi] \triangleq E_\varphi[\ln \frac{\varphi}{\psi}] = \sum_{\omega_i \in \Omega} \varphi[\omega_i] \ln \frac{\varphi[\omega_i]}{\psi[\omega_i]}. \quad \square$$

Relative entropy is, for a variety of reasons detailed in [2, ch.2], a very natural measure of discrepancy to use between a distribution and an approximation to it. Furthermore, and in contrast to $L_2$, it behaves very reasonably with respect to conditioning:

**Fact 1** For any $t$, $E_{\rho^{(t)}}[D[\mathcal{O}_{r_h}[\sigma^{(\bullet t)}] \| \mathcal{O}_{r_h}[\hat{\sigma}^{(t\bullet)}]]] \leq D[\sigma^{(\bullet t)} \| \hat{\sigma}^{(\bullet t)}]$, where $\rho^{(t)} = (\sigma^{(\bullet t)} \mathcal{O})$ is the prior on the response at time $t$.

Unfortunately, we seem to have simply shifted the problem from one place to another. While relative entropy is better behaved with respect to $\mathcal{O}$, there is no known contraction result for $\mathcal{T}$. Indeed, until now, the only related properties that seem to have been known [2, p.34] are that relative entropy never increases by transition through a stochastic process (i.e., that $D[\mathcal{T}[\sigma^{(t\bullet)}] \| \mathcal{T}[\hat{\sigma}^{(t\bullet)}]] \leq D[\sigma^{(t\bullet)} \| \hat{\sigma}^{(t\bullet)}]$), and that it ultimately tends to zero for a very broad class of processes (i.e., $D[\mathcal{T}^k[\sigma^{(t\bullet)}] \| \mathcal{T}^k[\hat{\sigma}^{(t\bullet)}]] \to 0$ as $k \to \infty$ when $\mathcal{T}$ is ergodic). Unfortunately, we need much stronger results if we wish to bound the accumulation of the error over time. One of the main contributions of this paper is the first proof (to our knowledge) that a stochastic transition contracts relative entropy at a geometric rate.

We now prove that a stochastic process $\mathcal{T}$ does, indeed, lead to a contraction in the relative entropy distance. It will be useful later on (for the case of DBNs) to consider a somewhat more general setting, where the sets of states 'before' and 'after' the stochastic transition are not necessarily the same. Thus, let $\Omega = \{\omega_1, ..., \omega_n\}$ be our *anterior* state space, and $\Omega' = \{\omega'_1, ..., \omega'_{n'}\}$ be our *ulterior* state space. Let $\mathcal{Q}$ be an $n \times n'$ stochastic matrix, representing a random process from $\Omega$ to $\Omega'$. Let $\varphi$ and $\psi$ be two given anterior distributions over $\Omega$, and let $\varphi'$ and $\psi'$ be the corresponding ulterior distributions induced over $\Omega'$ by $\mathcal{Q}$.

Our goal is to measure the minimal extent to which the process $\mathcal{Q}$ causes the two distributions to become the same. In the worst case, there is no common starting point at all: all of the mass in one distribution $\varphi$ is on some state $\omega_{i_1}$ while all of the mass in the other distribution $\psi$ is on some other state $\omega_{i_2}$. However, the stochastic nature of the process causes each of them to place some weight on any posterior state $\omega'_j$: the probability $\varphi'[\omega'_j]$ is $\mathcal{Q}[\omega'_j \mid \omega_{i_1}]$ while $\psi'[\omega'_j]$ is $\mathcal{Q}[\omega'_j \mid \omega_{i_2}]$. Thus, while none of the probability mass of $\varphi$ and $\psi$ was in agreement, $\varphi'$ and $\psi'$ must agree on $\omega'_j$ for a mass of $min[\mathcal{Q}[\omega'_j \mid \omega_{i_1}], \mathcal{Q}[\omega'_j \mid \omega_{i_2}]]$. Based on this insight, we have the following natural characterization of the mixing properties of our process:

**Definition 3** For a Markov process with stochastic transition model $\mathcal{Q}$, the *minimal mixing rate* of $\mathcal{Q}$ is

$$\gamma_\mathcal{Q} \triangleq min_{i_1,i_2} \sum_{j=1}^{n'} min[\mathcal{Q}[\omega'_j \mid \omega_{i_1}], \mathcal{Q}[\omega'_j \mid \omega_{i_2}]]. \quad \square$$

We will show that stochastic propagation by $\mathcal{Q}$ is guaranteed to reduce relative entropy by a factor of $1 - \gamma_\mathcal{Q}$. The proof relies on the following lemma, which allows us to isolate the probability mass in the two distributions which is guaranteed to mix:

**Lemma 2** *For any $\gamma \leq \gamma_\mathcal{Q}$ and any given $\varphi$ and $\psi$, the matrix $\mathcal{Q}$ admits an additive contraction decomposition $\mathcal{Q} = \mathcal{Q}^\Gamma + \mathcal{Q}^\Delta$, where $\mathcal{Q}^\Gamma$ and $\mathcal{Q}^\Delta$ are non-negative matrices such that, for all $i$, $\sum_{j=1}^{n'} \mathcal{Q}^\Gamma_{i,j} = \gamma$, and for all $j$, $\sum_{i=1}^{n} \varphi[\omega_i] \mathcal{Q}^\Gamma_{i,j} = \sum_{i=1}^{n} \psi[\omega_i] \mathcal{Q}^\Gamma_{i,j}$.*

Intuitively, we have argued that, for at least a certain portion of their probability mass, $\varphi'$ and $\psi'$ must agree. The matrix $\mathcal{Q}^\Gamma$ captures that "portion" of the transition that unifies the two. The two conditions state that this unification must happen with probability at least $\gamma_\mathcal{Q}$, and that the two distributions must, indeed, agree on on that portion.

Based on this lemma, our contraction result now follows easily. Essentially, the argument is based on a construction that makes explicit the different behavior of the process corresponding to the two parts of the contraction decomposition $\mathcal{Q}^\Gamma, \mathcal{Q}^\Delta$. As suggested on Figure 2, we split the process into two separate phases: in the first, the process "decides" whether to contract, and in the second the appropriate transition occurs. That is, we define a new intermediate state space $\Omega^\dagger$, which contains a new distinguished *contraction* state $c$ and otherwise is identical to $\Omega$. We separate out the cases in which the process is guaranteed to contract $\varphi$ and $\psi$ by having them correspond to an explicit transition to $c$. From $c$, the two processes behave identically, according to $\mathcal{Q}^\Gamma$, while from the remaining states in $\Omega^\dagger$, they behave according to $\mathcal{Q}^\Delta$. We are now in a position to prove our contraction theorem:

**Theorem 3** *For $\mathcal{Q}$, $\varphi$, $\psi$, $\varphi'$, $\psi'$ as above:*

$$D[\varphi' \| \psi'] \leq (1 - \gamma_\mathcal{Q}) D[\varphi \| \psi].$$

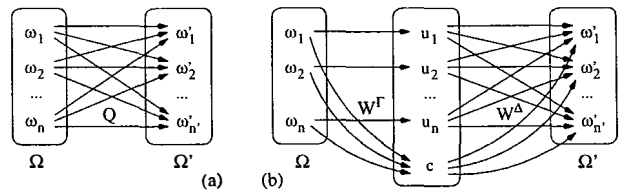

Figure 2: Decomposition used in Theorem 3: (a) generic Markov transition process; (b) two-stage process equivalent to it for $\varphi$ and $\psi$. Here, arrows denote stochastic transitions between states.

**Proof** Fix $\varphi$ and $\psi$, and define a new two-phase process with one Markovian transition $\mathcal{W}^\Gamma$ from $\Omega$ to $\Omega^\dagger$ and another $\mathcal{W}^\Delta$ from $\Omega^\dagger$ to $\Omega'$, where $\Omega^\dagger = \{u_1, ..., u_n, c\}$ is a new state set (see Figure 2). Intuitively, the state $u_i$ corresponds to $\omega_i$ while the state $c$ corresponds to the state obtained if the process contracts. The process $\mathcal{W}^\Gamma$ transitions to the contraction state with probability $\gamma$ and preserves its state with probability $1 - \gamma$; i.e., $\mathcal{W}^\Gamma[\omega_i \leadsto c] = \gamma$ while $\mathcal{W}^\Gamma[\omega_i \leadsto u_i] = 1 - \gamma$. The process $\mathcal{W}^\Delta$ behaves like $\mathcal{Q}^\Delta$ from the states $u_i$; from $c$, it duplicates the aggregate behavior of $\varphi$ on $\mathcal{Q}^\Gamma$, i.e., $\mathcal{W}^\Delta[u_i \leadsto \omega'_j] = \mathcal{Q}^\Delta_{i,j}/(1 - \gamma)$ while $\mathcal{W}^\Delta[c, \omega'_j] = \sum_i \varphi[\omega_i] \mathcal{Q}^\Gamma_{i,j}/\gamma$.

It is easy to show that the decomposed process is equivalent to the original process $\mathcal{Q}$ *for $\varphi$ and $\psi$* (this will not be the case for other distributions). To show that $D[\varphi' \| \psi'] \leq (1 - \gamma)D[\varphi \| \psi]$, we note that $D[\varphi^\dagger \| \psi^\dagger] = (1 - \gamma)D[\varphi \| \psi]$. As $\mathcal{W}^\Delta$ is Markovian, we also have $D[\varphi' \| \psi'] \leq D[\varphi^\dagger \| \psi^\dagger]$, hence the claim. ∎

## 4  Compound processes

Our results so far apply to any discrete state Markovian process. However, their potential impact is greatest in cases where an explicit representation of the belief state is infeasible. DBNs, which allow very complex processes to be specified compactly, often fall into this category. How well do our results apply to DBNs? Unfortunately, the answer is somewhat discouraging, even for very highly structured DBNs. As an extreme example, imagine a process composed of $N$ binary variables evolving independently, flipping their value each time slice with probability $\delta$. Each variable, viewed as a separate Markov process, has a mixing rate of $2\delta$. Thus, one may expect the mixing rate of the process as a whole to be as good; indeed, since all of the processes are independent, we could hardly expect otherwise.

However Theorem 3 tells a different story: computing $\gamma$ for the transition matrix of the compound process as a whole, one gets a discouragingly small value which is $\leq (4\delta)^{N/2}$. Is our definition of the mixing rate simply too pessimistic? Unfortunately not. The fallacy is in our assumption that local mixing properties would automatically carry over to the compound process. Each subprocess is rapidly mixing for belief states over its own variable only. If the belief state of the compound process involves dependencies between variables belonging to different subprocesses, then our contraction rate can, indeed, be as bad as our prediction. Assume, for example, that the true distribution $\varphi$ gives probability 1 to the state (i.e., assignment of values to variables) $\langle 0, ..., 0 \rangle$ while the approximate distribution $\psi$ gives probability $p$ to that state and $1 - p$ to its opposite $\langle 1, ..., 1 \rangle$. We can view the state space as a hypercube, and each of these distributions as a mass assignment to vertices of the hypercube. A single step through the transition matrix "diffuses" the mass of the two distributions randomly around their starting points. However, the probability that the diffusion process around one point reaches the other is exponentially low, since all bits have to flip in one or the other of the two distributions.

One approach to improving our results for decomposed processes is to make some additional assumptions about the structure of the belief state distributions, e.g., that they decompose. Clearly, such an assumption would be unreasonable for the true belief state, as its lack of decomposability was the basis for our entire paper. Somewhat surprisingly, it suffices to make a decomposability assumption only for the *approximate* belief state. That is, we can show that if the process decomposes well, and the estimated belief state decomposes in a way that matches the structure of the process, then we get significantly better bounds on the error contraction coefficient, regardless of the true belief state. Thus, as far as error contraction goes, the properties of the true belief state are not crucial: only the approximate belief state and the process itself. This is very fortunate, as it is feasible to enforce decomposability properties on the approximate belief state, whereas we have no control over the true belief state.

Formally, it is most convenient to describe our results in the framework of factored HMMs [17]; in the next section, we discuss how they can be applied to dynamic Bayesian networks. We assume that our system is composed of several subprocesses $T_l$. Each subprocess has a state with a Markovian evolution model. The state of subprocess $l$ at time $t$ is written $S_l^{(t)}$. The evolution model $T_l$ is a stochastic mapping from the states of some set of processes at time $t$ to the state of process $l$ at time $t + 1$. We say that subprocess $l$ *depends on* subprocess $l'$ if $T_l$ depends on the value of $S_{l'}^{(t)}$. Our model also allows a set of response variables at each time $t$, which can depend arbitrarily on the states of the processes at time $t$; however, as we are interested primarily in the contraction properties of our system, the properties of the response variables are irrelevant to our current analysis.

We begin by considering the simple case where our subprocesses are completely independent, i.e., where subprocess $l$ depends only on subprocess $l$. We also assume that our approximate belief state decomposes along the same lines. Clearly, this case is not interesting in itself: if our subprocesses are really independent, our belief state would never become correlated in the first place, and we would not need to approximate it. However, this analysis lays the foundation for the general case described below.

**Theorem 4** *Let $T_1, ..., T_L$ be $L$ independent subprocesses, let $\gamma_l$ be the mixing rate of $T_l$, and let $\gamma = \min_l \gamma_l$. Let $\varphi$ and $\psi$ be distributions over $S_1^{(t)}, ..., S_L^{(t)}$, and assume that the $S_l^{(t)}$ are marginally independent in $\psi$. Then $D[\varphi' \| \psi'] \leq (1 - \gamma) D[\varphi \| \psi]$.*

Thus, if we have a set of independent subprocesses, each of which contracts, *and* our approximate belief state decomposes along the same lines as the process, then the contraction of the process as a whole is no worse than the contraction of the individual subprocesses. Since each subprocess involves a much smaller number of states, its transition probabilities are likely to be reasonably large (assuming it is stochastic enough). This analysis usually results in a much better mixing rate.

Now consider the more general case where the processes interact. Assume that subprocess $l$ depends on subprocesses $l_1, ..., l_k$. Then, $T_l$ defines a probability $P(S_l \mid S_{l_1}, ..., S_{l_k})$. This transition probability can be defined as a transition matrix, but one whose anterior and ulterior state spaces can be different. Luckily, in Section 3, we allowed for exactly this possibility, so that the mixing rate of $T_l$ is well-defined. Let $\gamma_l$ be the mixing rate of $T_l$, and let $\gamma = \min_l \gamma_l$. If our approximate belief state, as before, respects the process structure, then we can place a bound on the mixing rate of the entire process. This bound depends both on $\gamma$ and on the process structure, and is given by the following theorem.

**Theorem 5** *Consider a system consisting of $L$ subprocesses $T_1, ..., T_L$, and assume that each subprocess depends on at most $r$ others; each subprocess influences at most $q$ others; and each $T_l$ has minimum mixing rate $\gamma_l \geq \gamma$. Let $\varphi$ and $\psi$ be distributions over $S_1^{(t)}, ..., S_L^{(t)}$, where the $S_l^{(t)}$ are independent in $\psi$. Then: Then $D[\varphi' \| \psi'] \leq (1 - \gamma^*) D[\varphi \| \psi]$, where $\gamma^* = (\gamma/r)^q$.*

**Proof idea** We illustrate the basic construction for a simple example; generalization to arbitrary structures is straightforward. Consider two processes $T_X$ and $T_Y$ as above, and assume that $T_Y$ depends on $T_X$. Our basic construction follows the lines of the proof of Theorem 3: we split the transition of each process into two phases where the first chooses whether or not to contract and the second concludes the transition in a way that depends on whether the process has contracted. Note, however, that the variable $X$ plays a role both in $T_X$ and in $T_Y$, and that the transitions of these two subprocesses are conditionally independent given $X$. Thus, $X$ cannot make a single decision to contract and apply it in the context of both processes. We therefore introduce two separate intermediate variables for $T_X$, $X_X^\dagger$ and $X_Y^\dagger$, where the first decides whether $X$ contracts in the context of $T_X$ and the second whether it contracts in the context of $T_Y$. Our decomposition is as follows:

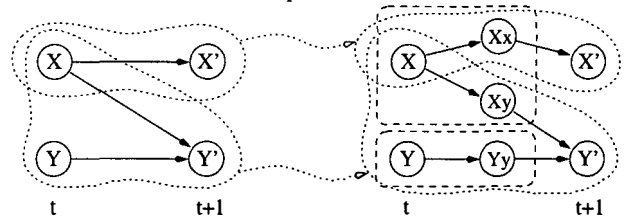

Our transitions for the partitioned model are defined so as to make sure that the parts of the two-phase processes marked with the dotted lines induce the same behavior as their corresponding one-phase processes, *for the distributions $\varphi[X]$ and $\psi[X]$*. (As in the proof of Theorem 3, this equivalence only holds for this pair of distributions.) The construction is essentially identical to that of Theorem 3, except that that the transition from $X$ to the distinguished contraction state $c_X^X$ of $X_X^\dagger$ is taken with some probability $\lambda_1$ to be specified below.

To get equivalence for the $\mathcal{T}_Y$ process, we must deal with a slight subtlety: The transition from $X_Y^\dagger, Y^\dagger$ to $Y'$ must behave as if the first phase contracted if *either* of the two antecendents $X_Y^\dagger, Y^\dagger$ is in the contraction state. (There is no choice, because if even one is in the contraction state, the process no longer has enough information to transition according to $\mathcal{Q}_Y^\Delta$.) Thus, if $X \to X_Y^\dagger$ contracts with probability $\lambda_2$ and $Y \to Y^\dagger$ with probability $\lambda_3$, the process as a whole contracts with probability $1 - (1 - \lambda_2)(1 - \lambda_3)$. Therefore, in order for us to be able to construct a contraction decomposition $\mathcal{Q}_Y^\Gamma, \mathcal{Q}_Y^\Delta$, we must select $\lambda_1, \lambda_2, \lambda_3$ so as to satisfy $\lambda_1 \leq \gamma_{\mathcal{Q}_X}$ and $1 - (1 - \lambda_2)(1 - \lambda_3) \leq \gamma_{\mathcal{Q}_Y}$. Assuming that $\gamma_{\mathcal{Q}_X} = \gamma_{\mathcal{Q}_Y} = \gamma$, we have that $\lambda_2 = \lambda_3 = 1 - \sqrt{1 - \gamma}$ is one legitimate selection; observe that $1 - \sqrt{1 - \gamma} \geq \gamma/2$.

To analyze the actual contraction rate of the process as a whole, we first analyze the contraction from the initial variables $X, Y$ to the intermediate variables $X_X^\dagger, X_Y^\dagger, Y^\dagger$; the contraction for the two phases is no smaller. Our analysis here uses a somewhat different process structure—the one shown using the dashed lines—than the one used to show the correctness of the partition. These processes—$\mathcal{W}_X$ from $X$ to $\{X_X^\dagger, X_Y^\dagger\}$ and $\mathcal{W}_Y$ from $Y$ to $Y^\dagger$—are independent (by design). Furthermore, we have assumed that $X$ and $Y$ are independent in $\psi$. Thus, the conditions of Theorem 4 apply[1] and the contraction of the process from $X, Y$ to $X_X^\dagger, X_Y^\dagger, Y^\dagger$ is the minimum of the contraction of $\mathcal{W}_X$ and of $\mathcal{W}_Y$. Straightforwardly, the contraction of $\mathcal{W}_Y$ is $\lambda_3$. However, $\mathcal{W}_X$ contracts—loses all information about its original state—only when both $X_X$ and $X_Y$ enter their contraction state. These events are independent, hence the probability that they both occur is the product $\lambda_1 \lambda_2$.

In general, if a subprocess $\mathcal{Q}_l$ has a contraction ratio $\gamma_l$ and depends on $r_l$ subprocesses, we have to choose the contraction factor $\lambda$ for each influencing process (assuming for simplicity that all are chosen to be equal) to be $1 - \sqrt[r_l]{1 - \gamma_l} \geq \gamma/r$, resulting in a linear reduction in the contraction rate. On the other hand, each influence of a subprocess $\mathcal{Q}_l$ on another subprocess involves the construction of another intermediate variable, each of which contracts independently. The total contraction of $\mathcal{W}_{X_l}$ is the product of the individual contractions, which is $\gamma^* = \lambda^q$, an exponential reduction. Putting it all together, the overall contraction rate admits the lower bound $\gamma^* = (\gamma/r)^q$. ∎

We see that interconnectivity between the processes costs us in terms of our contraction ratio. The cost of "incoming" connectivity is a linear reduction in the contraction rate whereas the cost of "outgoing" connectivity is exponential. Intuitively, this phenomenon makes sense: if a process influences many others, it is much less likely that its value will be lost via contraction.

---

[1] The theorem was stated for processes where the anterior and ulterior state space is identical, but the same proof applies to the more general case.

## 5 Efficient monitoring

As we suggested throughout the paper, one of the main applications of our results is to the task of monitoring a complex process. As we mentioned, the exact monitoring task is often intractable, requiring that we maintain a very large belief state. We proposed an alternative approach, where we maintain instead an *approximate* belief state. As we now show, our contraction results allow us to bound the cumulative error arising from such an approximate monitoring process. We also investigate one particular application of this approach, in the contex of DBNs.

### 5.1 Approximate monitoring

Recall the notation of Section 2 about belief states, and denote by $\tilde{\sigma}^{(t)}$ our compactly represented approximate belief state at time $t$. The choice of compact representation depends on the process. If, for example, our process is composed of some number of weakly interacting subprocesses—e.g., several cars on a freeway—it may be reasonable to represent our belief state at a given time using our marginal beliefs about its parts (e.g., individual vehicles). In a hybrid process, as we said, we may want to use a fixed-size mixture of Gaussians.

The approximate belief state $\tilde{\sigma}^{(t)}$ is updated using the same process as $\sigma^{(t\bullet)}$: we propagate it through the transition model, obtaining $\hat{\sigma}^{(\bullet t+1)}$, and condition on the current response, obtaining $\hat{\sigma}^{(t+1\bullet)}$. However, $\hat{\sigma}^{(t+1\bullet)}$ does not usually fall into the same family of compactly-represented distributions to which we chose to restrict our belief states. In order to maintain the feasibility of our update process, we must approximate $\hat{\sigma}^{(t+1\bullet)}$, typically by finding a "nearby" distribution that admits a compact representation; the result is our new approximate belief state $\tilde{\sigma}^{(t+1)}$. In our freeway domain, for example, we may compute our new beliefs about the state of each vehicle by projecting $\hat{\sigma}^{(t+1\bullet)}$, and use the cross product of these individual belief states as our approximation; the resulting distribution is the closest (in terms of relative-entropy) to $\hat{\sigma}^{(t+1\bullet)}$. In our continuous process, we could project back into our space of allowable belief states by approximating the distribution using a fixed number of Gaussians.

We begin by analyzing the error resulting from this type of strategy, i.e., the distance between the true posterior belief state $\sigma^{(t+1\bullet)}$ and our approximation to it $\tilde{\sigma}^{(t+1)}$. Intuitively, this error results from two sources: the "old" error which we "inherited" from the previous approximation $\tilde{\sigma}^{(t)}$, and the "new" error derived from approximating $\hat{\sigma}^{(t+1\bullet)}$ using $\tilde{\sigma}^{(t+1)}$. Suppose that each approximation introduces an error of $\varepsilon$, increasing the distance between the exact belief state and our approximation to it. However, the contraction resulting from the state transitions serves to drive them closer to each other, reducing the effect of old errors by a factor of $\gamma$. The various observations move the two even closer to each other on expectation (averaged over the different possible responses). Therefore, the expected error accumulated up to time $t$ would behave as $\varepsilon + (1 - \gamma)\varepsilon + ... + (1 - \gamma)^{t-1}\varepsilon \leq \varepsilon \sum_i (1 - \gamma)^i = \varepsilon/\gamma$.

To formalize this result, we first need to quantify the error resulting from our approximation. Our new approximate belief state $\bar{\sigma}^{(t)}$ is an approximation to $\hat{\sigma}^{(t\bullet)}$. Most obviously, we would define the error of the approximation as the relative entropy distance between them—$D[\hat{\sigma}^{(t\bullet)} \| \bar{\sigma}^{(t)}]$. However, our error is measured relative to $\sigma^{(t\bullet)}$ and not to $\hat{\sigma}^{(t\bullet)}$. Therefore, we use the following definition:

**Definition 4** We say that an approximation $\bar{\sigma}^{(t)}$ of $\hat{\sigma}^{(t\bullet)}$ incurs error $\varepsilon$ relative to $\sigma^{(t\bullet)}$ if

$$D[\sigma^{(t\bullet)} \| \bar{\sigma}^{(t)}] - D[\sigma^{(t\bullet)} \| \hat{\sigma}^{(t\bullet)}] \leq \varepsilon. \quad \square$$

Our last theorem now follows easily by induction on $t$:

**Theorem 6** Let $\mathcal{T}$ be a stochastic process whose mixing rate is $\gamma$, assume that we have an approximation scheme that, at each phase $t$, incurs error $\varepsilon$ relative to $\sigma^{(t\bullet)}$. Then for any $t$, we have:

$$E_{\rho(1,\ldots,t)}[D[\sigma^{(t\bullet)} \| \bar{\sigma}^{(t)}]] \leq \varepsilon/\gamma,$$

where the expectation is taken over the possible response sequences $r_{h_1}, \ldots, r_{h_t}$, with the probability ascribed to them by the process $\mathcal{T}$.

Of course, it is not trivial to show that a particular approximation scheme will satisfy the accuracy requirement of $\varepsilon$, essentially because its definition involves the true belief $\sigma^{(t\bullet)}$, which is usually not known. Nevertheless, a sufficient condition for this requirement is $\max_i \ln[\hat{\sigma}^{(t\bullet)}[s_i]]/\bar{\sigma}^{(t)}[s_i] \leq \varepsilon$; and the left-hand side quantity—the maximum log relative error caused by the approximation scheme at time $t$—is often easy to assess for a given approximation step.

Let us understand the guarantees provided by this theorem. First, the bound involves relative entropy between the two entire belief states. In certain applications, we may be interested in errors for individual variables or for subsets of variables. Fortunately, any bound on the entire distribution immediately carries over to any projection onto a subset of variables [2]. Furthermore, we note that bounds on relative entropy immediately imply bounds on the $L_1$ error, as $\|\varphi - \psi\|_1 \leq (2 \ln 2 D[\varphi \| \psi])^{1/2}$. Second, note that the bounds are on the expected error; the error for specific sequences of evidence are much weaker. In particular, the error after a very unlikely sequence of evidence might be quite large. Fortunately, our contraction result holds for arbitrary distributions, no matter how far apart. Thus, even if momentarily $\sigma^{(t\bullet)}$ and $\bar{\sigma}^{(t)}$ are very different, the contraction property will reduce this error exponentially.

### 5.2 Monitoring in DBNs

We now consider a specific application of this general approach. We consider a process described as a DBN, and a factored belief state representation (where certain subsets of variables are forced to be independent). In this case, the process is specified in terms of an ordered set of state variables $X_1, \ldots, X_n$. The probability model of a DBN is typically described using a *2-TBN* (a two time-slice temporal Bayesian network), as shown in Figure 1. The 2-TBN associates each variable with a conditional probability distribution $P[X_k^{(t+1)} \mid Parents(X_k^{(t+1)})]$, where $Parents(X_k^{(t+1)})$ can contain any variable at time $t$ and such variables at time $t + 1$ that preceed $X_k$ in the total ordering. This model represents the conditional distribution over the state at time $t + 1$ given the variables at time $t$. Let $C$ be the set $\{X : X^{(t)} \in \cup_k Parents(X_k^{(t+1)})\}$.

To capture the idea of a subprocess, we partition the set $C$ into disjoint subsets $X_1, \ldots, X_L$. Our partition must satisfy the requirement that no $X_l$ may be affected by another $X_{l'}$ within the same time slice; i.e., if $X \in X_l$, then $X^{(t+1)}$ cannot have as an ancestor a variable $Y^{(t+1)}$ for $Y \in X_{l'} \neq X_l$. Note that a time slice may also contain non-persistent variables, e.g., sensor readings; but since none of them may ever be an ancestor of a canonical variable, we allow them to depend on any persistent variable in their time slice. Since the various clusters $X_l$ correspond to our subprocesses from Theorem 5, we shall maintain an approximate belief state in which the $X_l$ are independent, as prescribed.

The approximate monitoring procedure for DBNs follows the same lines as the general procedure described in Section 2: At each point in time, we have an approximate belief state $\bar{\sigma}^{(t)}$, in which the $X_l$ are all independent. We propagate $\bar{\sigma}^{(t)}$ through the transition model, and then condition the result on our observations at time $t+1$. We then compute $\bar{\sigma}^{(t+1)}$ by projecting $\hat{\sigma}^{(t+1\bullet)}$ onto each $X_l$; i.e., we define $\bar{\sigma}^{(t+1)}[X_l] = \hat{\sigma}^{(t+1\bullet)}[X_l]$, and the entire distribution as a product of these factors.

In the case of DBNs, we can actually accomplish this update procedure quite efficiently. We first generate a clique tree [13] in which, for every $l$, some clique contains $X_l^{(t)}$ and some clique contains $X_l^{(t+1)}$. A standard clique tree propagation algorithm can then be used to compute the posterior distribution over every clique. Once that is done, the distribution over $X_l^{(t+1)}$ is easily extracted from the appropriate clique. Further savings can be obtained if we assume a stationary DBN and a static approximation scheme.

In order to apply this generic procedure to a particular problem, we must define a partition of the canonical variables, i.e., choose a partition of the process into subprocesses. Our analysis in the previous sections can be used to evaluate the alternatives. The tradeoffs, however, are subtle: Subprocesses with a small number of state variables allow more efficient inference. They also have a smaller transition matrix and therefore their mixing rate is likely to be better. On the other hand, our subprocesses need to be large enough so that there are no edges between subprocesses within a single time slice. Furthermore, making our subprocesses too small increases the error incurred by the approximation of assuming them to be independent. Specifically, if we have two (sets of) variables that are highly correlated, splitting them into two separate subprocesses is not a good idea. Our experimental results illustrate these tradeoffs.

### 5.3 Experimental results

We validated this algorithm in the context of two real-life DBNs: the WATER network [8], used for monitoring the biological processes of a water purification plant; and the BAT network [6], used for monitoring freeway traffic (see Figure 3). We added a few evidence nodes to WATER, which did not have any; these duplicate a few of the state variables with added noise. The experimentation methodology is as follows: starting from the same initial prior (the uniform prior), we monitor the process evolution using our approximate method with some fixed decomposition, and compare it at every $t$ to the exact inference, which is emulated by using our algorithm with the trivial partition (all canonical state variables in a single cluster). Observations are simulated by sampling the evidence variables according to the exact distribution.

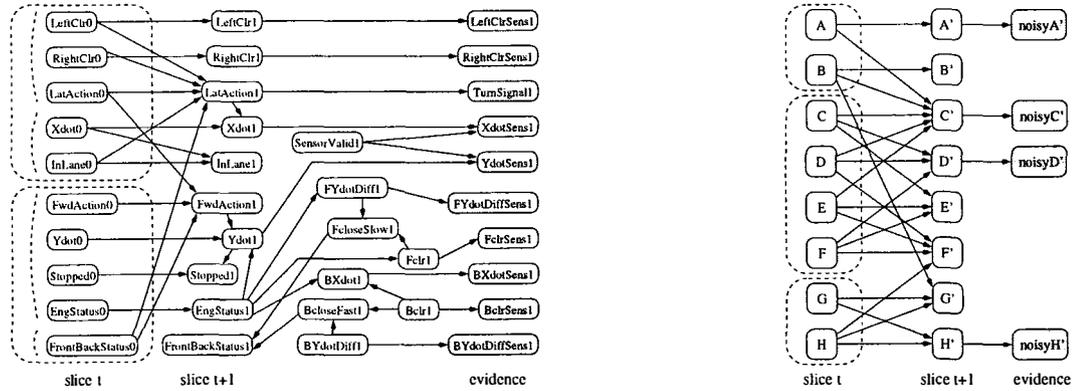

Figure 3: The canonical form 2-TBNs for the DBNs used in our experiments: (a) the BAT network; (b) the WATER network. The dotted lines indicate some of the clusterings used in our approximations.

Figure 4(a) shows the evolution of relative entropy for the BAT network on a typical run, using the shadowed nodes as evidence nodes. In this network, the belief state consists of 10 variables roughly partitioned in two weakly interacting groups of 5: we choose this "5+5" partition for our approximation scheme. On an UltraSparc 2, the approximate monitoring took about 0.11 seconds per time slice, as compared to 1.72 for the exact inference, yielding a 15-fold speedup. In terms of accuracy, the error averages at 0.0007, remaining very low most of the time with, as we would expect, a few sparsely distributed spikes to somewhat larger values, peaking at 0.065. We also note that it does not appear to grow over the length of the run, as predicted by our analysis. Since in practical applications the emphasis is often on a few selected variables, we also computed the $L_1$ errors for the beliefs over the two variables 'LateralAction' and 'ForwardAction' (i.e., the belief states marginalized over each of them). Their qualitative pattern was roughly similar to Figure 4(a), they respectively averaged 0.00013 and 0.0019, and remained bounded by 0.02 and 0.07 over the 1000-step run of our experiment.

Similar tests were conducted on the WATER network shown on Figure 3(b), and using a decomposed belief over the 3 clusters A-B, C-D-E-F, and G-H. Over a run of length 3000, the error remained bounded by 0.06 with the exception of one outlier to 0.14, and averaged 0.006 over that run. Running times were 0.19 sec/slice for the approximation, vs. 6.02 sec/slice for the reference (a 31-fold speedup).

To investigate the effect of our approximate belief state representation, we ran the following experiment on three different projections: the "5+5" clustering already introduced, a "3+2+4+1" clustering obtained by further breaking down the 5-clusters (see Figure 3(a)), and a "3+3+4" clustering chosen irrespectively of the network topology.[2]

First, we used our theoretical tools to predict how well each of these would perform, compared with each other. The two determining factors here are the stepwise approximation error $\varepsilon$ and the overall mixing rate bounded by $\gamma^*$. The former is directly linked to the expressiveness of the approximation; so according to this criterion we have in decreasing order of quality: "5+5" $\gg$ "3+3+4" $\succeq$ "3+2+4+1". To assess the mixing rates, we computed the vector $\vec{\gamma}$ of mixing rates for all the clusters, and used it together with the topological quantities $q$ and $r$ to calculate $\gamma^*$ from Theorem 5. For the "5+5" clustering, this gave $\vec{\gamma} = \langle 0.00040, 0.0081 \rangle$, $r = 2, q = 2$, and thus $\gamma^* = (\gamma_{min}/r)^q = 4 \times 10^{-8}$. For the "3+2+4+1" clustering, $\vec{\gamma} = \langle 0.00077, 0.080, 0.0081, 0.96 \rangle$, $r = 3, q = 2$, so $\gamma^* = 7 \times 10^{-8}$. For the "3+3+4" clustering, $\vec{\gamma} = \langle 0.0022, 0.020, 0.0034 \rangle$, $r = 3, q = 3$, so $\gamma^* = 4 \times 10^{-10}$. For this criterion, we get "3+2+4+1" $\simeq$ "5+5" $\gg$ "3+3+4": the latter is heavily penalized by the higher inter-connectivity of its clusters.

The second step was to compare this with the actual monitoring accuracy of our algorithm in each case. The results are shown in Figure 4(b) (averaged over 8 different runs and plotted on logarithmic scale). The "5+5" clustering (lower curve) has an average error of 0.0006, its error is always lower than the "3+2+4+1" clustering (medium curve) for which the error averages 0.015. Both are clearly superior to the "3+3+4" clustering, for which the average error reaches 0.13.[3] We observe that the errors obtained in practice is significantly lower than that predicted by the theoretical analysis. However, qualitatively, the behaviour of the different clusterings is consistent with the predictions of the theory. In particular, the quality of a clustering is very sensitive to its topology, as expected.

Interestingly, the accuracy can be further improved by using *conditionally* independent approximations. In the WATER network, for example, using a belief state decomposed into the overlapping clusters A-B-C-D-E, C-D-E-F-G, G-H yields, for the same sequence of observations as above, an average error of just 0.0015. Also, the error now remains bounded by 0.018 throughout, reducing the maximum error by a factor of 8. Approximate inference here took 0.47 sec/slice, a 13-fold speedup over exact inference.

---

[2]Strictly speaking, the "3+2+4+1" and "3+3+4" clusterings contain clusters that are not subprocesses according to our definition, as there are edges between clusters within the same time slice. However, it is impossible to find smaller clusterings in the BAT network that satisfy this restriction, and the empirical results still serve to illustrate the main points.

[3]The speedups relative to exact inference were respectively 15, 20, 20 for the "5+5", "3+2+4+1", "3+3+4" clusterings.

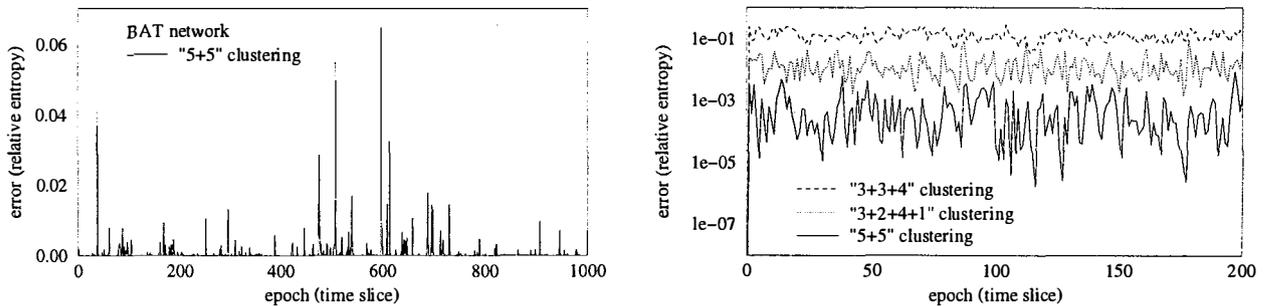

Figure 4: Experimental results: (a) relative entropy error for a typical run using the BAT network; (b) comparison of relative entropy error for three different approximate belief state representations (BAT network, curves averaged over 8 runs).

We also tested the effect of evidence on the quality of our approximation. Without evidence, the error curve is very smooth, and converges relatively soon to a constant error, corresponding to the error between the correct stationary distribution of the process and our approximation to it. With evidence, the error curve has a much higher variance, but its average is typically much lower, indicating that evidence further boosts contraction (as opposed to merely being harmless). The effect of evidence was more beneficial with better clusterings.

## 6 Conclusion and extensions

We investigated the effect of approximate inference in a stochastic process. We showed that the stochastic nature of the process tends to make errors resulting from an approximation disappear rapidly as the process continues to evolve. Our analysis relies on a novel and important theorem proving a contraction result for relative entropy over stochastic processes. We also provided a more refined analysis for processes composed of weakly interacting subprocesses.

We applied the idea of approximate inference to the monitoring task. Exact inference for this task is infeasible in complex stochastic processes with a large number of states. Even if the process is highly structured, exact inference (e.g., [11]) is forced into intractability by the full correlation of the belief state that invariably occurs. We propose an approach whereby the algorithm maintains an approximate belief state with compact representation. This belief state is propagated forward from one time slice to the next, and the result is approximated so as to maintain a compact representation. Our analysis guarantees that the errors from our approximations do not accumulate.

This approach provides a general framework for approximate inference in stochastic processes. We also presented one instantiation of this general framework to the class of structured processes composed of weakly interacting subprocesses. For this case, we provide a more refined analysis which results in significantly lower error bounds. We also presented experimental results showing that our approach works extremely well for real-life processes. Indeed, we get order of magnitude savings even for small processes, at the cost of a a very low error in our approximation. For larger processes, we expect the savings to be much greater. Both our theoretical and our empirical results clearly show that our ability to execute accurate effective inference for a stochastic process depends heavily on its structure. Thus, our result is the first to show how the structure of a stochastic process can be exploited for inference.

There has been fairly little work on approximate inference in complex temporal models.[4] The early work of [14] considers a simple approach of using domain knowledge to simply eliminate some of the variables from each time slice. The random sampling approach of [10] can also be viewed as maintaining an approximate belief state, albeit one represented very naively as a set of weighted samples. The recent work of [12] extends this idea, using the random samples at each time slice as data in order to *learn* an approximate belief state. Both of these ideas can be viewed as falling into the framework described in this paper. However, neither contains any analysis nor an explicit connection to the structure of the process.

The recent work of [7] and [16] utilize mean field approximation in the context of various types of structured HMMs. Of these approaches, [7] is the closest to our work (to the part of it dealing with structured processes). In their approach, the compound process is approximated as being composed of independent subprocesses, whose parameters are chosen in a way that depends on the evidence. This approach differs from ours in several ways. First, it applies only to situations where the subprocesses are, in fact, independent, and only become correlated due to observations. Our work applies to much richer models of partially decomposed systems. Second, their approximation scheme is based on making the *subprocesses* independent throughout their length, whereas our approach accounts for the correlations between the subprocesses at each time slice, and then marginalizes them out. Finally, their analysis proves that their approximation is the closest possible to the true distribution *among the class of distributions represented by independent processes*; our analysis, by contrast, provides explicit bounds (in expectation) on the total error.

---

[4]Some approximate inference algorithms for nontemporal Bayesian networks can also be applied to this task. Specifically, the *mini-bucket* approach of [5] is somewhat related to ours, as it also uses a (different) form of factoring decomposition during the course of inference. However, none of the potentially relevant algorithms are associated with any performance guarantees on the error of approximation over time.

One important direction in which our results can be extended relates to different representations of the belief state and of the process. Our current analysis requires that the belief state representation make the states of the subprocesses completely independent. Clearly, in many situations a more refined belief state representation is more appropriate. For example, as our experiments show, it can be very beneficial to make the states of two processes in our approximate belief state *conditionally independent* given a third; we would like to derive the formal conditions under which this is indeed an advantage. The results of [12] also demonstrate that other representations of a belief state might be useful, specifically ones that allow some context-sensitivity of the structure of the distribution. They also show that it is beneficial to allow the structure of the belief state representation to vary over time, thus making it more appropriate to the specific properties of the current belief state. In general, our basic contraction result applies to any approximation scheme for belief state representation; we intend to experiment with different alternatives and evaluate their contraction properties. Finally, we would like to apply our analysis to other types of processes requiring a compact representation of the belief state, e.g., processes with a continuous state space.

Another priority is to improve our analysis of the effect of evidence. Our current results only show that the evidence does not hurt too much. Our experiments, however, show (as we would expect) that the evidence can significantly reduce the overall error of our approximation.

There are many other tasks to which our techniques can be applied. For example, our analysis applies as is to the task of predicting the future evolution of the system. It also applies to tasks where the transition model depends on some action taken by an agent. Our contraction analysis is done on a phase-by-phase basis, and therefore does not require that the transition model be the same at each time slice. Thus, so long as the transition associated with each action is sufficiently stochastic, we can use our technique to predict the effects of a plan or a policy. A more speculative idea is to the task of constructing policies for POMDPs; there, a policy is a mapping from belief states to actions. Perhaps our more compact belief state representation will turn out to be a better basis for representing a policy.

We hope to generalize our contraction result to the case of reasoning backwards in time, allowing us to apply approximate inference to the task of computing our beliefs about a time slice given both past and future evidence. This inference task is the crucial component in algorithms that learn probabilistic models of stochastic processes from data. The learning process requires many iterations of this inference task, so that any improvement in efficiency could have major impact. Furthermore, by showing that the influence of approximations in the future cannot significantly affect our beliefs about the past, we could obtain theoretical support for the simple and natural algorithm that computes beliefs about a time slice from a fairly small window on both sides. The applications of this idea to online learning are clear.

## Acknowledgements

We gratefully acknowledge Eric Bauer, Lise Getoor, and Uri Lerner for work on the software used in the experiments, Raya Fratkina for help with the network files, and Tim Huang for providing us with the BAT network. Many thanks to Tom Cover, Nir Friedman, Leonid Gurvits, Alex Kozlov, Uri Lerner, and Stuart Russell for useful discussions and comments. This research was supported by ARO under the MURI program "Integrated Approach to Intelligent Systems", grant number DAAH04-96-1-0341, by DARPA contract DACA76-93-C-0025 under subcontract to Information Extraction and Transport, Inc., and through the generosity of the Powell Foundation and the Sloan Foundation.